\documentclass[journal]{IEEEtran}

\IEEEoverridecommandlockouts                              
\usepackage{graphics} 
\usepackage{epsfig} 
\usepackage{times} 
\usepackage{amsmath} 
\usepackage{amssymb}  
\usepackage{bm}
\usepackage[bookmarks=true]{hyperref}
\usepackage{xcolor}
\usepackage{color}
\usepackage[ruled,vlined]{algorithm2e}
\usepackage{balance}
\usepackage{array}
\usepackage{siunitx}  
\usepackage{booktabs}
\usepackage{multirow}
\usepackage{subcaption}
\usepackage{cite}
\usepackage{mwe} 
\usepackage{soul}
\usepackage{layout}

\definecolor{Gray}{gray}{0.9}
\definecolor{somegray}{rgb}{0.5, 0.5, 0.5}

\usepackage{etoolbox}

\let\oldtwocolumn\twocolumn
\renewcommand\twocolumn[1][]{%
    \oldtwocolumn[{#1}{
    \begin{center}
        \includegraphics[width=0.25\linewidth]{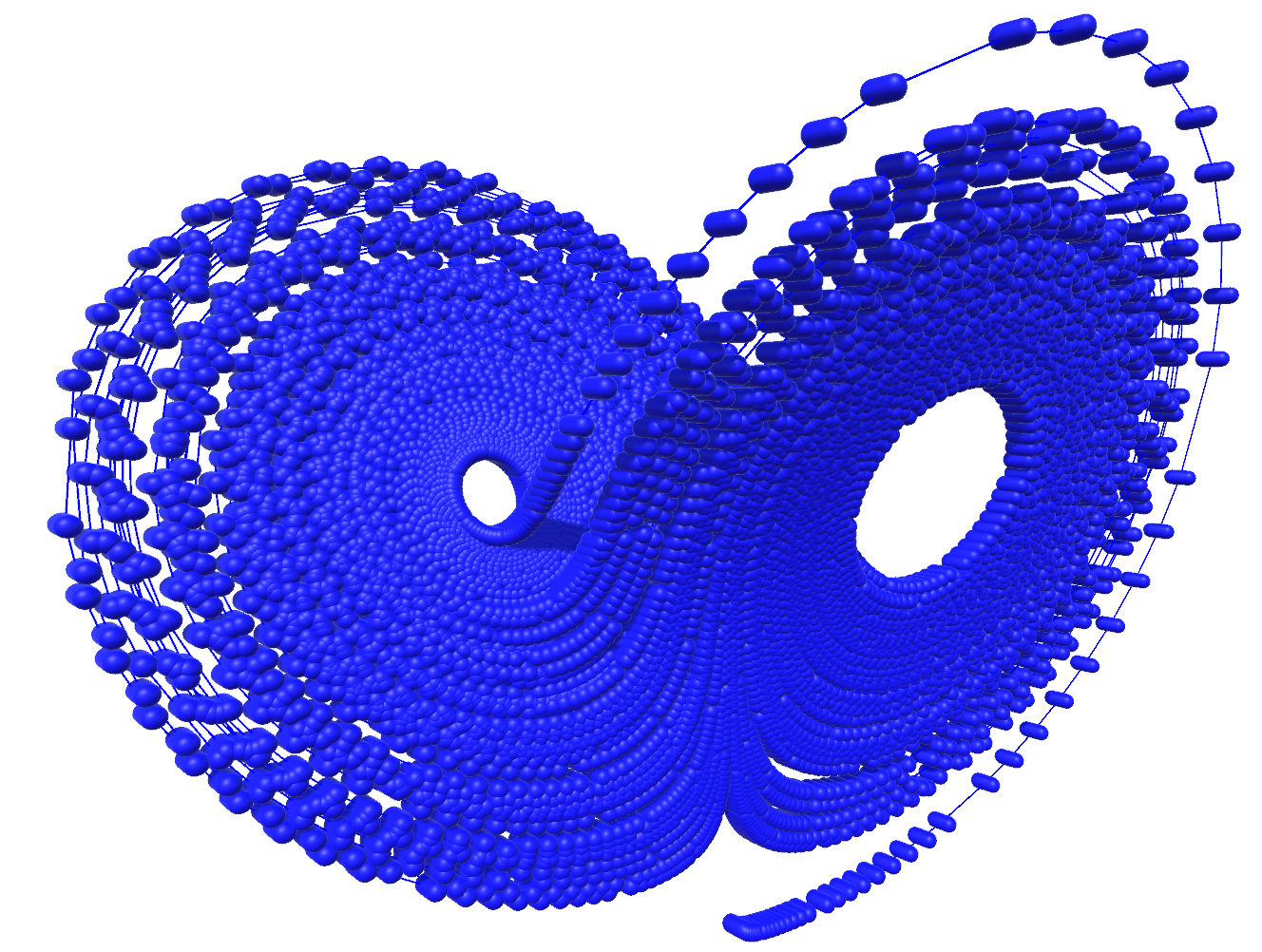}\hfill
        \includegraphics[width=0.25\linewidth]{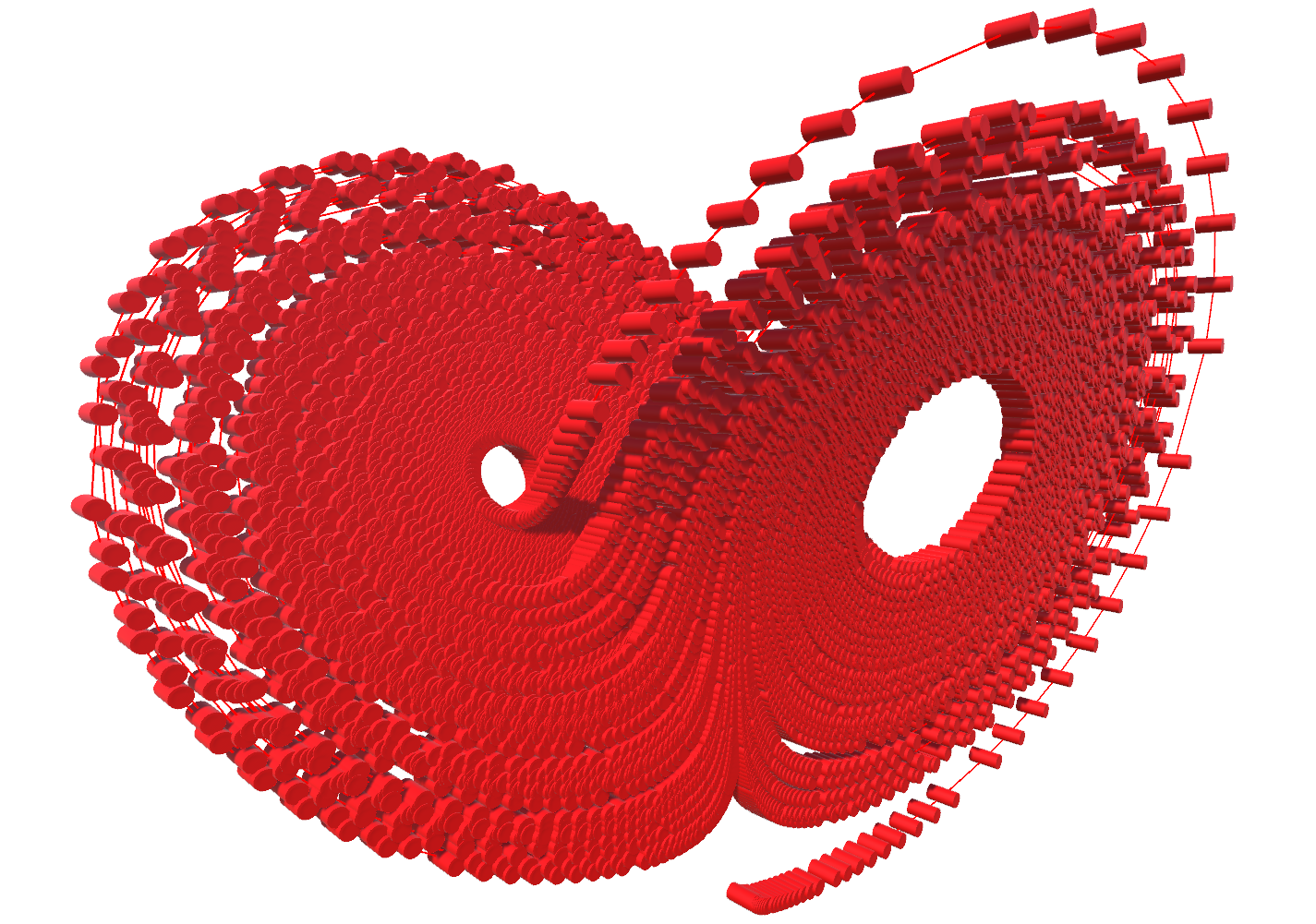}\hfill
        \includegraphics[width=0.25\linewidth]{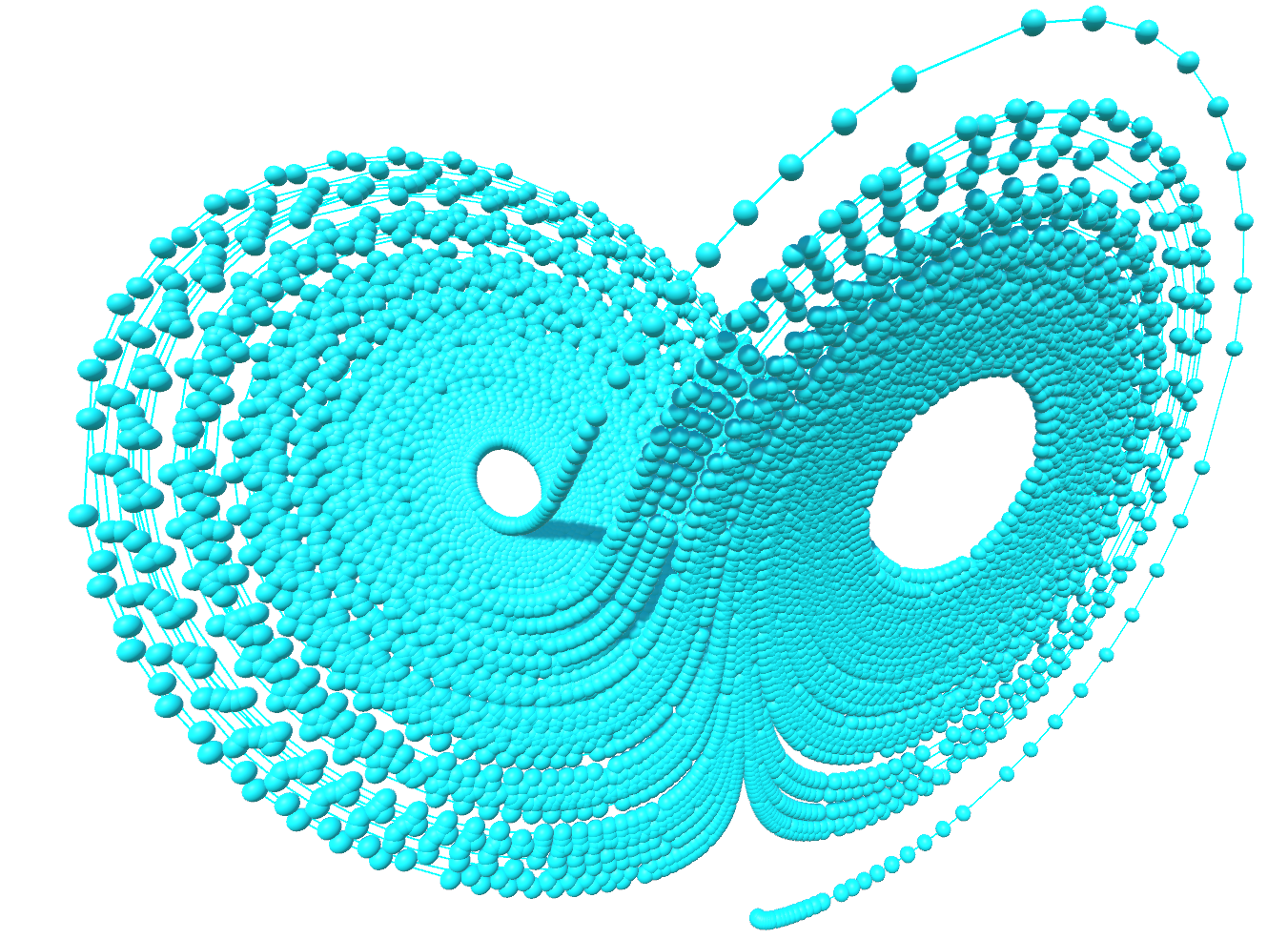}\hfill
        \includegraphics[width=0.25\linewidth]{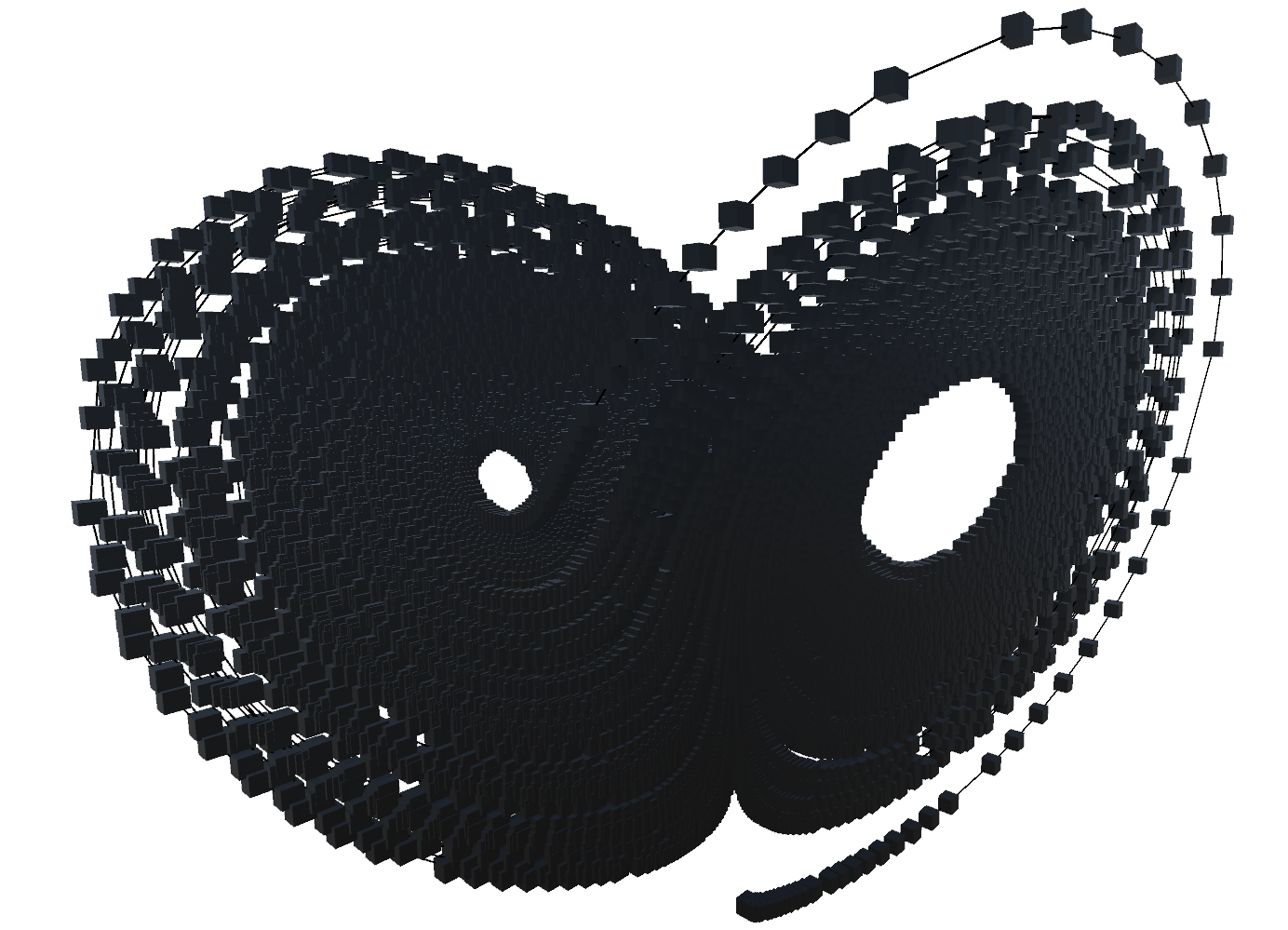}\hfill
        \newline
        \includegraphics[width=0.5\linewidth]{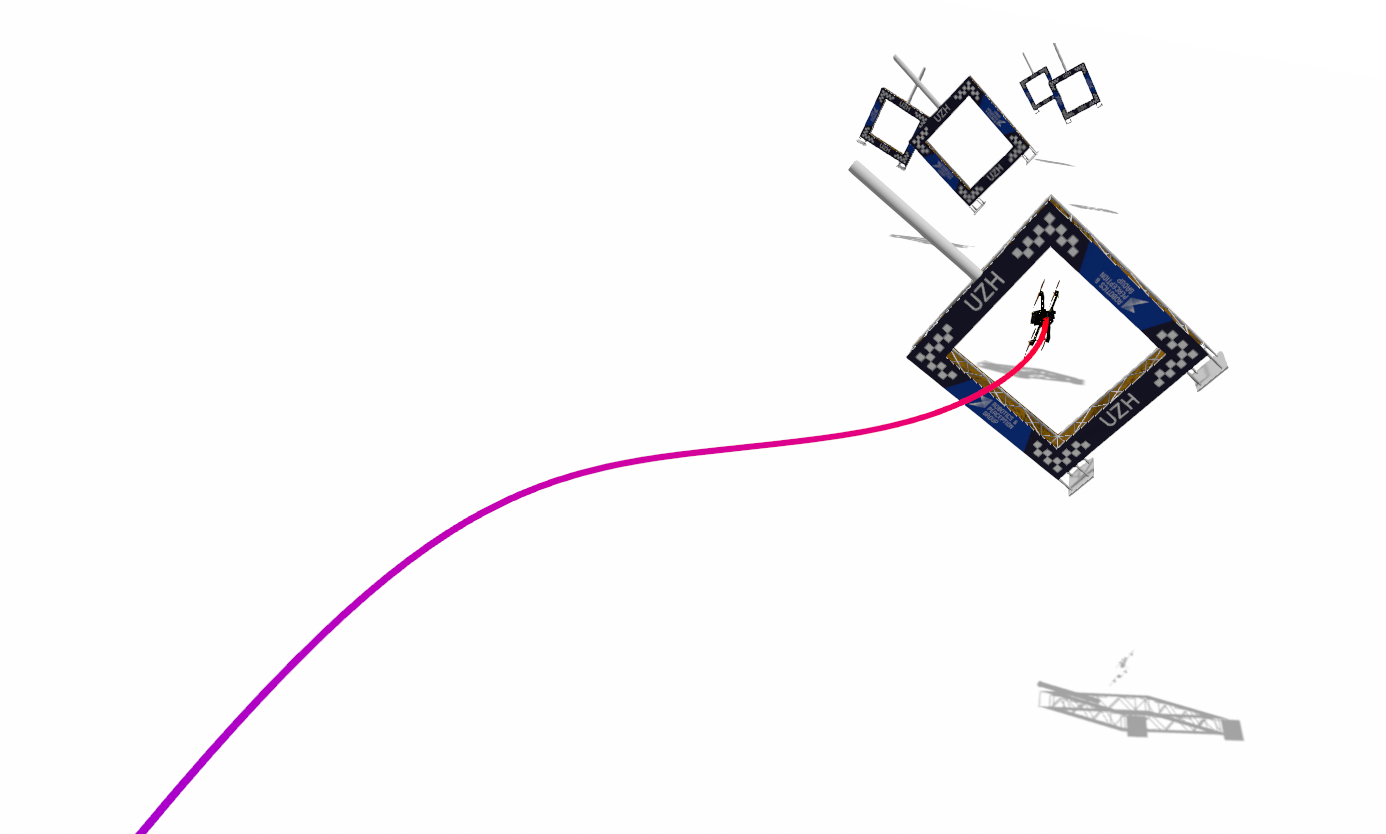}\hfill
        \includegraphics[width=0.5\linewidth]{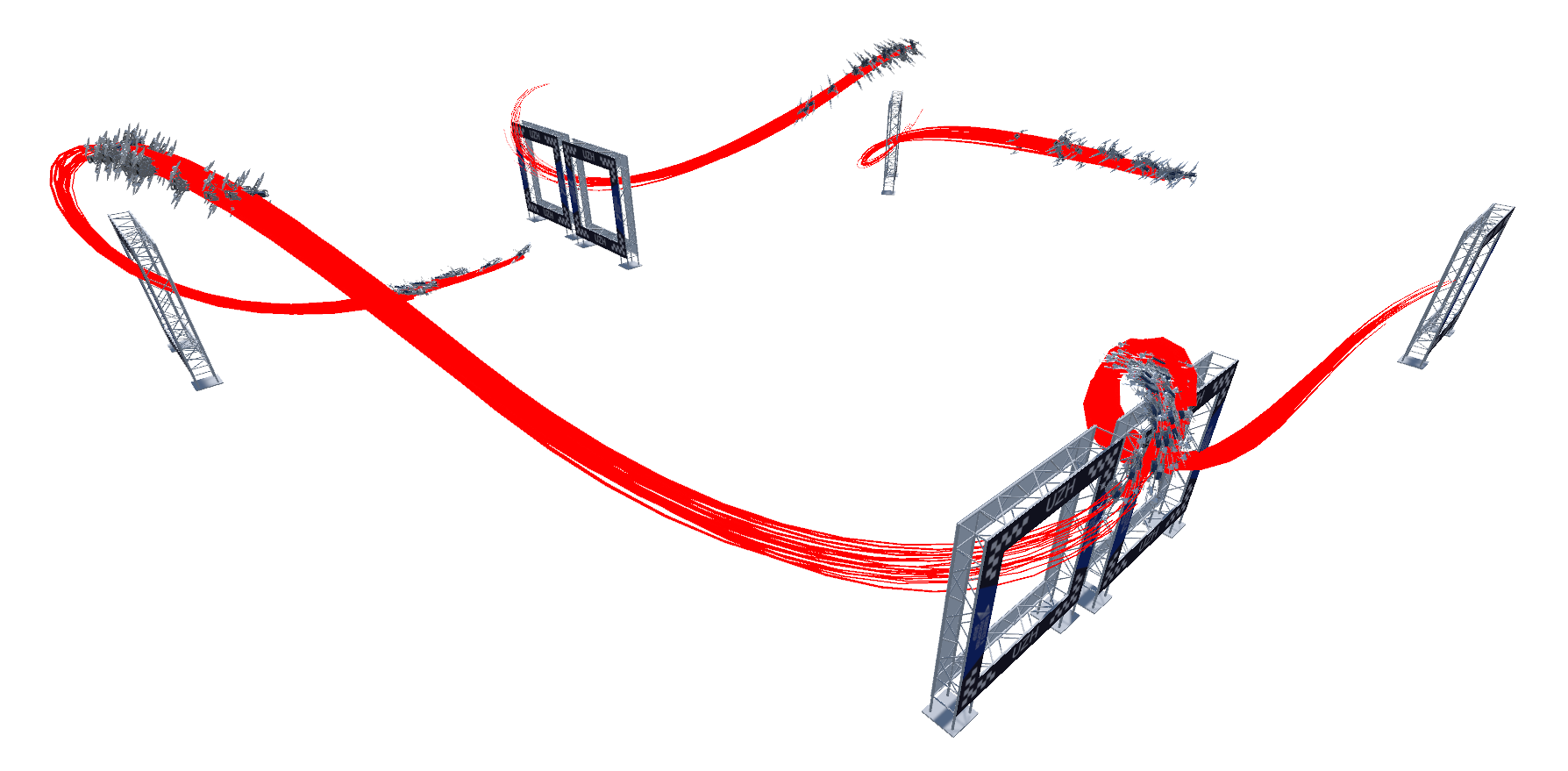}\hfill
      \captionof{figure}{Flymation is a flexible, interactive animation tool that allows real-time visualization and animation of 3D trajectories, with the capability of augmenting the trajectory using realistic 3D models, such as \emph{drone, cube, cylinder, and more.}  }
      \label{fig: teaser_img}
    \end{center}
    }
    ]
}

\title{ 
Flymation: \\ Interactive Animation for Flying Robots
}

\author{Yunlong Song, Davide Scaramuzza
    \thanks{The authors are with the Robotics and Perception Group, Department of Informatics, University of Zurich, and Department of Neuroinformatics, University of Zurich and ETH Zurich, Switzerland (\protect\url{http://rpg.ifi.uzh.ch}).
    This work was supported by the National Centre of Competence in Research (NCCR) Robotics through the Swiss National Science Foundation (SNSF) and the European Union’s Horizon 2020 Research and Innovation Programme under grant agreement No. 871479 (AERIAL-CORE) and the European Research Council (ERC) under grant agreement No. 864042 (AGILEFLIGHT).
    }
}

\begin{document}

\maketitle

\begin{abstract}
Trajectory visualization and animation play critical roles in robotics research. However, existing data visualization and animation tools often lack flexibility, scalability, and versatility, resulting in limited capability to fully explore and analyze flight data. To address this limitation, we introduce \emph{Flymation}, a new flight trajectory visualization and animation tool.
Built on the Unity3D engine, \emph{Flymation} is an intuitive and interactive tool that allows users to visualize and analyze flight data in real time. Users can import data from various sources, including flight simulators and real-world data, and create customized visualizations with high-quality rendering. With \emph{Flymation}, users can choose between trajectory snapshot and animation; both provide valuable insights into the behavior of the underlying autonomous system.
\emph{Flymation} represents an exciting step toward visualizing and interacting with large-scale data in robotics research.\\

\textbf{GitHub:} \url{https://github.com/uzh-rpg/flymation} 

\end{abstract}


\IEEEpeerreviewmaketitle

\section{Introduction}
\label{section: intro}
As autonomous algorithms evolve, robots are expected to execute increasingly complex tasks within highly complicated environments.
However, increasingly complicated tasks present a significant challenge to researchers as the amount of data generated scales exponentially. Data can provide valuable insights into a robot's behavior during research, development, and deployment. 
When visualizing or animating the data, such as a 3D trajectory, we can quickly identify patterns, trends, and relationships that would be much more difficult to discern from raw data. 
As such, a fundamental question is how researchers can effectively leverage existing data visualization tools to gain insights from this data.

Some of the most commonly used tools for data visualization include ROS (Robot Operating System), Matlab, Python, and R. While these tools offer flexibility for data visualization, they are also highly specialized and can be limited by their reliance on structured environments (such as installing MATLAB). This can create a significant barrier for individuals who may not have access to these specialized environments. Additionally, this limitation can hinder collaborations between professional researchers and individuals from other fields who may not be familiar with these tools.

This work explores using Unity3D for data visualization, which is a versatile game engine that can be used to create interactive visualizations for different tasks, thanks to its powerful photorealistic rendering capabilities. 
Unity3D has been mainly used for robotic simulation, such as Flightmare~\cite{song2020flightmare}, which is a flexible quadrotor simulator. 

This work focuses on \emph{interactive} data visualization and provides a new tool called \emph{Flymation}.
\emph{Flymation} offers a flexible and versatile solution for visualizing large-scale data for flying robots. \emph{Flymation} can visualize three-dimensional trajectory data augmented with photorealistic rendering. 
\emph{Flymation} is well-suited for both researchers and students since it is highly accessible. Particularly, it does not require any additional installation once a standalone application has been built, which means the application can be run directly on the target platform, such as Ubuntu or Windows.

\section{Related Work}
\label{section: relatedwork}

We review several existing tools that have been widely used by robotics and researchers. 
Our aim is to highlight the important features of each tool. Drawing inspiration from the successful work of our predecessors, we propose a paradigm shift and develop a new animation tool.
It is important to note that the choice of tool will ultimately depend on the specific needs of each project and the developer's preferences. 

\textbf{RViz}~\cite{rviz} is a 3D visualization tool that is widely used in robotics. It allows researchers to visualize robot models, sensor data, and other information in real time. RViz is part of the ROS (Robot Operating System) framework and is available for Linux, macOS, and Windows.

\textbf{Gazebo}~\cite{gazebo} is a free and open-source 3D simulation software for simulating robots, sensors, and environments. It can simulate complex robotic systems in various environments, supporting physics simulation, sensor models, and visualization. Gazebo can be integrated with ROS and other robotics frameworks.

\textbf{PlotJuggler}~\cite{plotjuggler} is a free and open-source data visualization tool that is specifically designed for time-series data. It is commonly used in robotics for visualizing data from sensors and other sources. PlotJuggler allows you to plot and compare multiple time-series data sources in real time. It includes features such as zooming, panning, and data filtering, which allow you to explore and analyze large datasets. It supports many data formats, including CSV and ROSBAG. PlotJuggler can be integrated with the ROS (Robot Operating System) framework.

\textbf{Matplotlib}~\cite{matplotlib} is a popular data visualization library in Python. It can be used to create 2D and 3D plots, histograms, and other types of visualizations. Matplotlib is widely used in robotics for visualizing data and generating high-quality figures for publications. 

\textbf{Plotly}~\cite{plotly} is an interactive data visualization library for Python and other programming languages. It can be used to create interactive 3D plots, heatmaps, and other visualizations. Plotly can be used in robotics for visualizing sensor data and other types of data.

\section{Methodology}
\label{section: method}

\begin{figure*}[!htp]
    \includegraphics[width=1\linewidth]{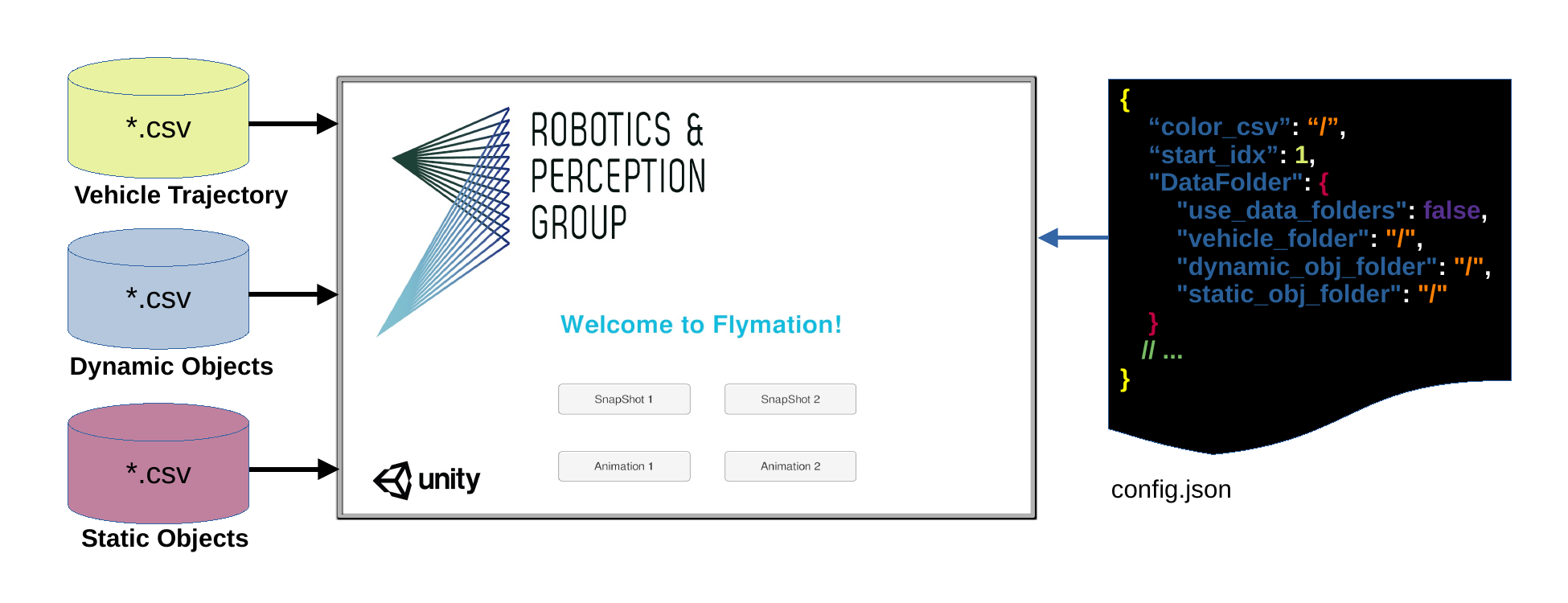}
  \captionof{figure}{Overview of the Flymation, a standalone application that allows interactive data visualization for flying robots. }
  \label{fig: method}
\end{figure*}

Flymation, as depicted in Fig.~\ref{fig: method}, is a flexible and versatile data visualization and animation framework.
Flymation is built on Unity3D (often referred to as ``Unity"), which is a popular cross-platform game engine for creating video games, virtual reality, simulation, and other interactive content. Unity offers powerful tools and features for creating 2D and 3D visualizations, including scripting, animation, and rendering.

\subsection{Implementation}
We implemented Flymation using the C\# programming language and the Unity Editor. In Unity development, scripting is a crucial concept. Unity provides a scripting API (Application Programming Interface) that enables developers to access and manipulate the engine's functionality through code.
With Unity scripting, we were able to load data from \emph{csv} files, create and customize 3D objects, generate line rendering, animate the trajectory, and more.
To promote collaboration and development, we have made our implementation open-source, allowing researchers to adapt our code to their own applications. We refer readers to our open-source implementation for more details. 

\subsection{Data Structure}

Flymation reads data at run time from three separate folders: a vehicle folder, a dynamic object folder, and a static object folder.
The vehicle folder contains a collection of \emph{csv} files that define the motion of a vehicle, such as a drone, through a sequence of state vectors. 
The dynamic object folder contains a collection of \emph{csv} files that define the motion of a rigid body, such as an obstacle.
Each state vector $\mathbf{h}_t = [\mathbf{p}_t, \mathbf{q}_t, \mathbf{v}_t, \mathbf{c}_t, \mathbf{s}_t]$ represents the position $\mathbf{p}_t \in \mathbb{R}^3$, orientation $\mathbf{q}_t \in \mathbb{R}^4$, linear velocity $\mathbf{v}_t \in \mathbb{R}^3$, color $\mathbf{c}_t \in \mathbb{R}^4$, and scale $\mathbf{s}_t \in \mathbb{R}^3$ of the rigid body at time step $t$. Hence, users can define the motion of a rigid body augmented with specific object types, such as cubes, with varying colors and sizes at different time steps.

Similarly, the static object folder contains files that specify the state of static objects. Each object state $\mathbf{h} = [\mathbf{p}, \mathbf{q}, \mathbf{c}, \mathbf{s}, \emph{obj}]$ comprises a position $\mathbf{p}$, orientation $\mathbf{q}$, color $\mathbf{c}$, scale $\mathbf{s}$, and object type $\emph{obj}$. This makes Flymation highly flexible, as objects can be easily configured with varying poses, colors, scales, and rendering representations at runtime.

\subsection{User Interface}
Flymation offers four user interfaces for users to interact with, including buttons, a \emph{json} script, a keyboard, and a mouse. 

The buttons serve as a graphical interface that allows users to easily select between visualizing a snapshot of the trajectory or animating it. When viewing a snapshot, users can choose between a continuous line representation or a time-lapse illustration, which is shown as a sequence of 3D virtual objects. Users can select between a static camera and a moving camera to animate the trajectory.

The \emph{json} script allows the user to specify the data's location and configure the color of a specific vehicle trajectory, which can be important when combining Flymation with other visualization tools, such as Matplotlib.  

The keyboard and mouse provide great flexibility for users to interact with Flymation.
For example, the user can use the mouse to zoom in or out of the view and rotate the trajectory in 3D. Using the keyboard, resetting the environment, and taking a screenshot of the view are easy.

\subsection{Snapshot and Animation}
Flymation offers users two different visualization functionalities: ``snapshot" and ``animation."

The ``snapshot" feature enables users to view the entire trajectory at once from any angle or position, with the possibility to augment the trajectory with different 3D objects and colors using Unity's realistic rendering capabilities. For instance, in Fig.~\ref{fig: teaser_img}, we showcase four butterfly-shaped trajectories generated via a Lorenzo attractor simulation with four different colors and four distinct objects.

On the other hand, the ``animation" feature displays an object's movement over time along a specific path. Users can customize the camera's position, whether it's at a fixed location or a third-person view that closely follows a moving object's motion. In Fig.~\ref{fig: teaser_img}, we demonstrate the ``animation" feature with 50 trajectories flown by 50 racing drones simultaneously in a realistic flying arena. The flow path's tail is highlighted in red, representing the trajectory's historical path.
\section{Applications}
\label{section: result}
This section highlights several applications~\cite{song2021autonomous, Yunlong2022policy, song2022learning} that have benefited from \emph{Flymation}.

\begin{figure*}[!htp]
    \includegraphics[width=0.33\linewidth]{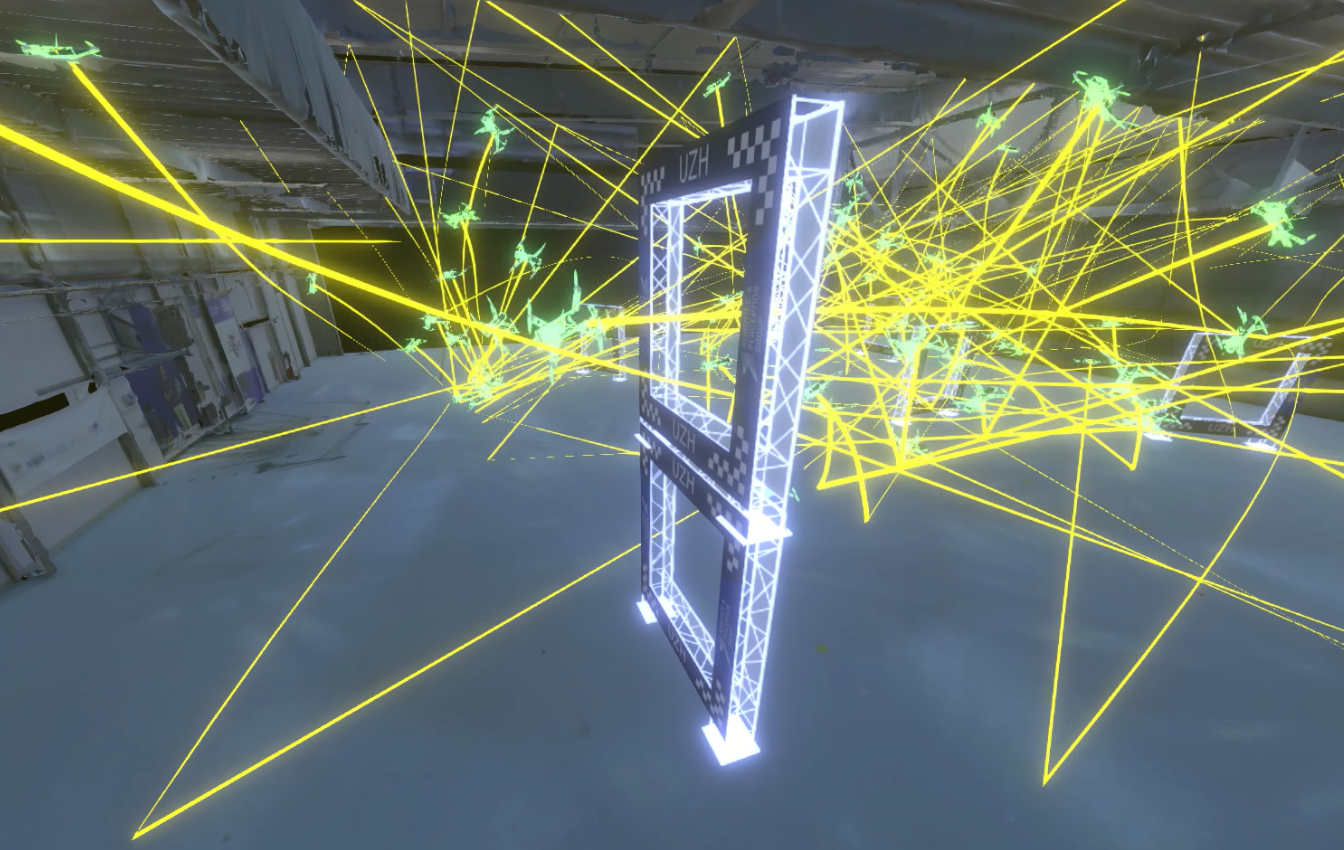}\hfill
    \includegraphics[width=0.33\linewidth]{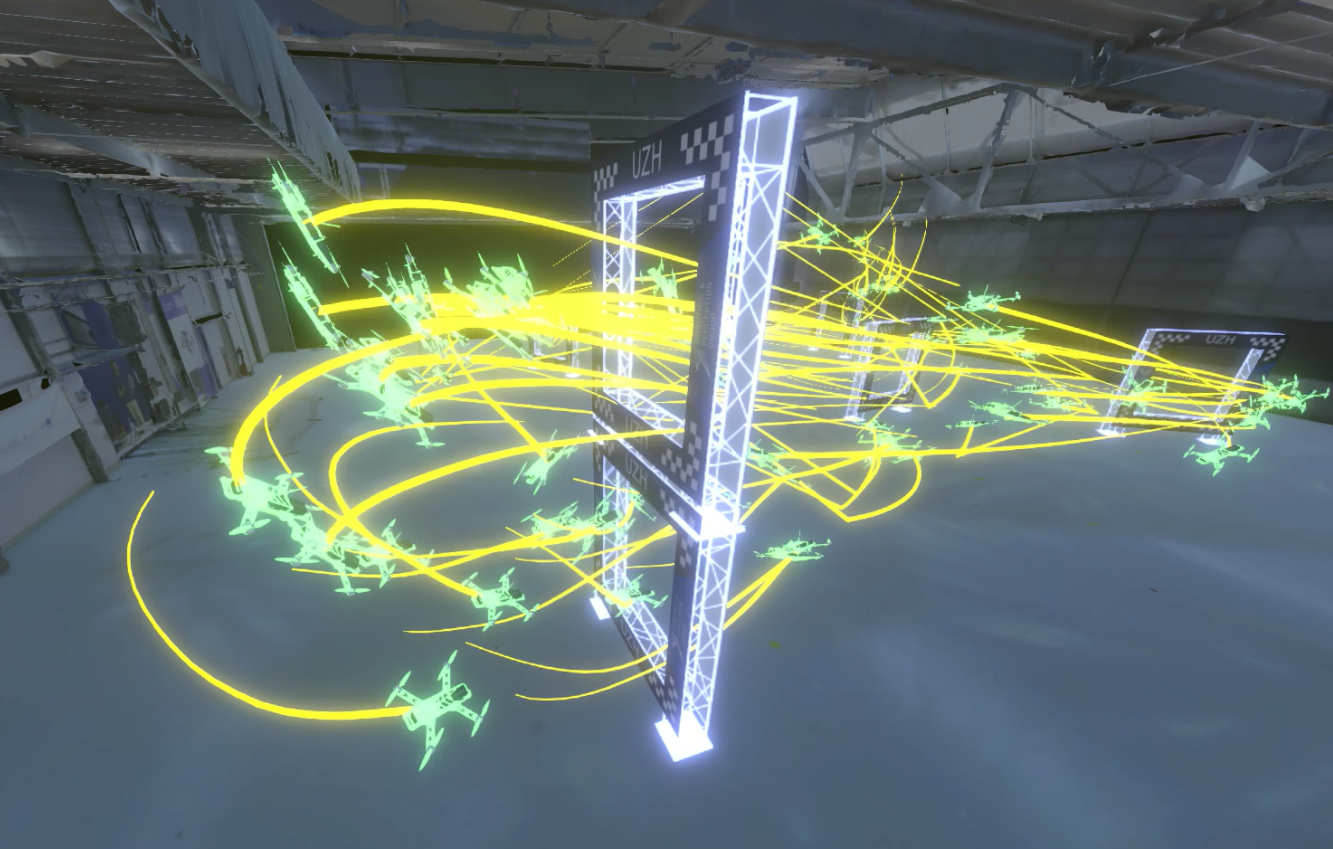}\hfill
    \includegraphics[width=0.32\linewidth]{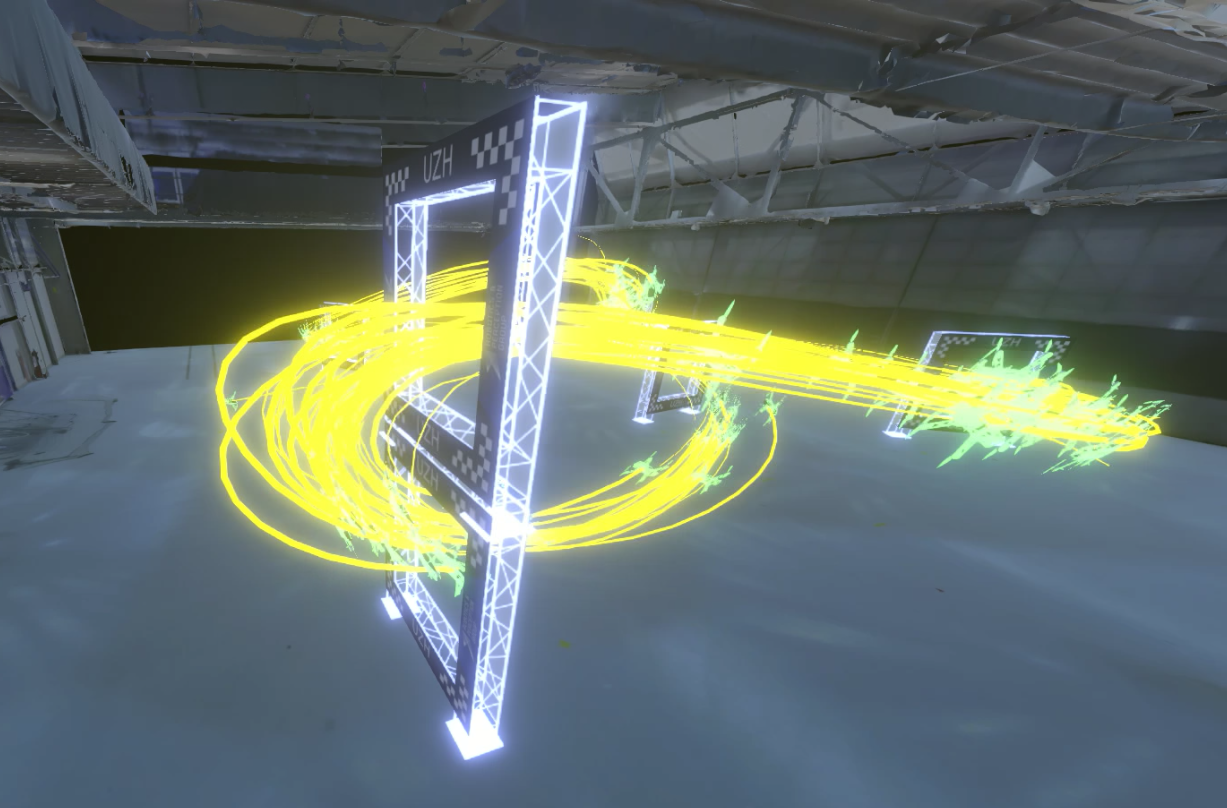}\hfill
  \captionof{figure}{Visualizing training data in reinforcement learning for drone racing. From left to right are sampled trajectories using policy trained at different iterations. }
  \label{fig: rl}
\end{figure*}

\subsection{Visualization in Reinforcement Learning}
Reinforcement learning (RL) is an approach that involves iterative trial-and-error interactions between an agent and the environment to learn an optimal control policy. However, applying RL to a specific task can be challenging and often requires tuning various aspects of the training pipeline. In addition to inspecting training curves, such as expected return, visualizing the agent's behavior during training can provide valuable insights into the learning progress and help accelerate algorithm design and tuning.

This section showcases the scalability and flexibility of Flymation in visualizing millions of data points in RL training~\cite{song2023reaching}. Specifically, we focus on the challenging task of autonomous drone racing. The objective of this task is to minimize the total lap time, which requires pushing the drone to its maximum limit. Our previous work has demonstrated the effectiveness of model-free RL in achieving this goal, but it requires millions of samples to learn an optimal policy. A fundamental question arises: how can such racing behaviors emerge from maximizing a single reward signal?

Figure~\ref{fig: rl} depicts three snapshots of the agent's behavior obtained from the sampled data. The first image illustrates the drone's random flying behavior with a randomly initialized neural network policy. The second image shows how the drone can successfully fly through gates at a relatively low speed after approximately 100 training iterations. Finally, after around 1000 training iterations (as shown in the last image), the policy can achieve optimal racing performance with high success rates and speed. We refer the reader to the attached video for an animation of the agent's behavior. 

\subsection{Visualization in Agile Drone Navigation}
Agile drone navigation in cluttered environments is an important problem since it opens up possibilities for many time-critical applications in the real world, such as research and rescue.  
In~\cite{penicka2022learning, song2022learning}, we leverage reinforcement learning and topological path planning to train robust neural network controllers for minimum-time quadrotor flight in cluttered environments.
The key ingredients of our approach to minimum-time flight in cluttered environments are three-fold: 1) generation of a
topological guiding path using a probabilistic roadmap, 2) a novel task formulation that combines progress maximization along the guiding path with obstacle avoidance, and 3) combining curriculum training with deep reinforcement learning.

\begin{figure}[!htp]
    \includegraphics[width=1.\linewidth]{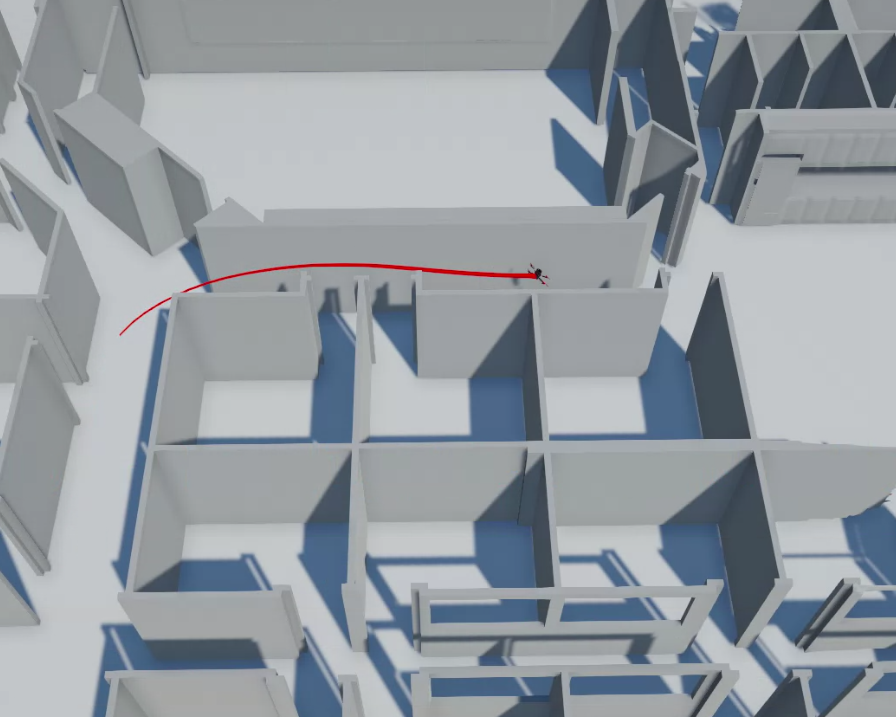} 
  \captionof{figure}{Visualizing drone navigation in a cluttered office.}
  \label{fig: office}
\end{figure}

Fig.~\ref{fig: office} shows a visualization of drone navigation in a cluttered office. 
We show that the presented method can achieve a high success rate in flying minimum-time policies in cluttered environments and outperforms classical approaches that rely on traditional planning and control. 
The policy is trained entirely in simulation and then transferred to the real world without fine-tuning, with
a peak flight speed of 42~\SI{}{\kilo\meter\per\hour} and a maximum accelerations of 3.6g.

\subsection{Visualization in Optimal Control}
Using optimal control in robot navigation typically involves dividing the task into planning and control. It's crucial that the planned trajectory is collision-free and that the controller can precisely track the trajectory to ensure the task's success. Flymation enables the visualization of both the planned trajectory and the trajectory tracking performance of the downstream controller.

\begin{figure}[!htp]
   \includegraphics[width=1.\linewidth]{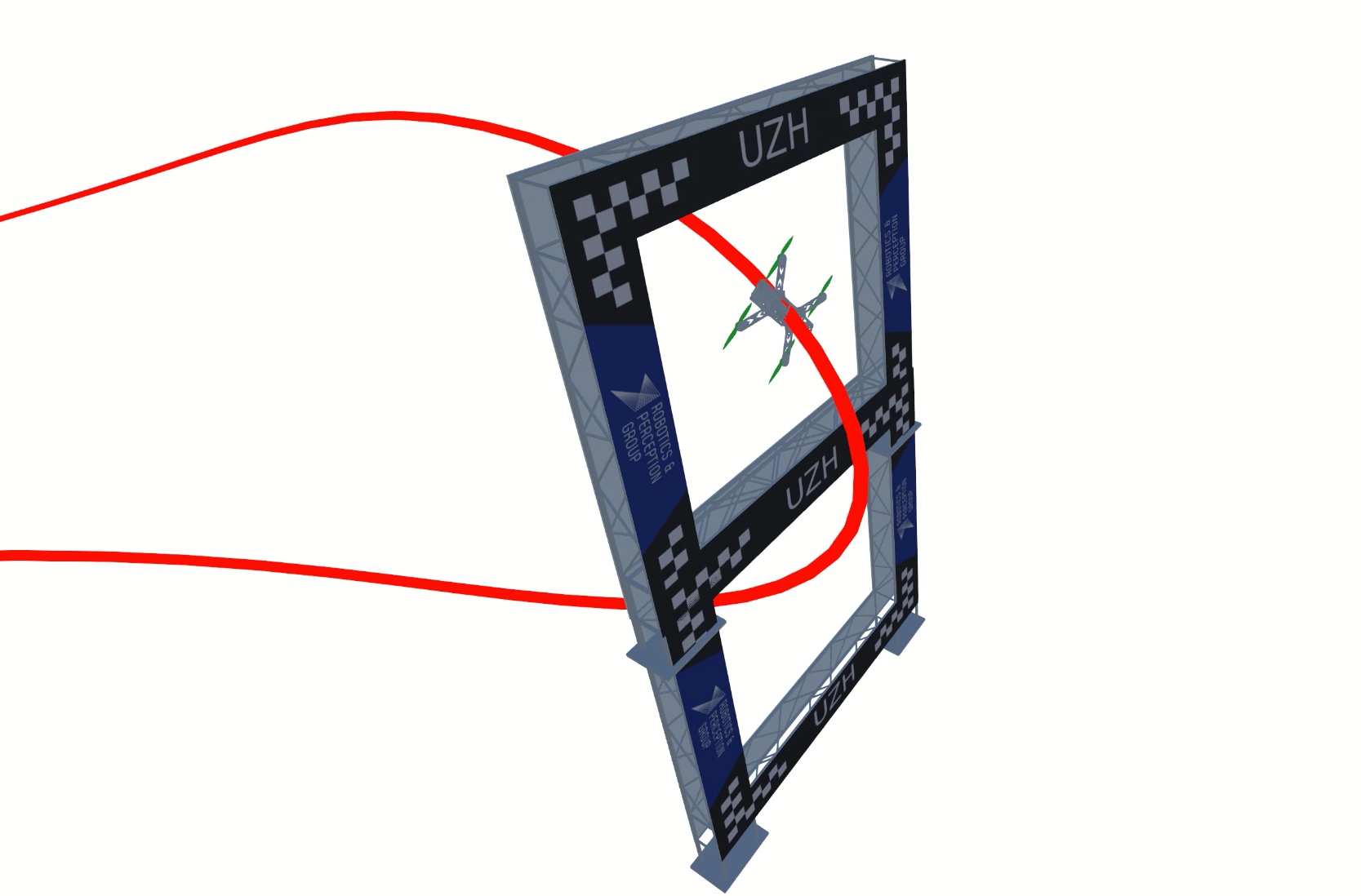} 
  \captionof{figure}{Visualizing trajectory tracking in optimal control.}
  \label{fig: tracking}
\end{figure}

Figure~\ref{fig: tracking} showcases tracking a planned trajectory using model predictive control in drone racing, highlighting a significant advantage of Flymation. This advantage stems from its ability to use a realistic prefab model for visualizing the drone's pose in a three-dimensional environment and its relationship with the environment. As a result, we can observe the drone's orientation at different parts of the race track. Additionally, the distance between the drone and the gate edges can be easily visualized, providing valuable insight for designing an autonomous system using optimal control.

\subsection{General-Purpose Trajectory Animation}
Flymation is a versatile 3D trajectory visualization and animation tool that is not limited to flying robots. 
Its flexibility is due to an important design choice: it accepts \emph{csv} files as input, which allows customizing the trajectory at run time with various configurations, such as color and 3D models. 

Fig.~\ref{fig: teaser_img} demonstrates Flymation's general-purpose trajectory visualization and animation capabilities using a Lorenz system's trajectory rendered with different colors and object types. Fig.~\ref{fig: office} demonstrated visualizing a quadrotor navigating in a cluttered environment at high speed.

\section{Conclusion}
\label{section: conclusion}
This work showcases the flexibility and scalability of using Unity for 3D trajectory visualization and animation. Our solution, Flymation, focuses on general rigid body representations, which are particularly well-suited for flying robots like drones. Compared to existing tools, Flymation offers several important advantages.
First, it provides user-friendly interfaces that allow users to interact with their data. Second, it allows for rich representations, ranging from simple spheres to complex and realistic 3D models. And third, once a Unity Standalone has been built, Flymation does not require installing any additional packages or coding, making it easy for professional robotic researchers to collaborate with individuals from other fields.

Overall, our work opens up exciting new possibilities for researchers to create and share interactive visualizations of robotic data. We see Flymation as an important step towards large-scale data visualization in the era of artificial intelligence and big data.
Extending Flymation to other robotic domains, such as humanoid robots and legged robots, represents a promising direction for future development.


\bibliographystyle{IEEEtran}
\bibliography{ref}

\end{document}